\PassOptionsToPackage{prologue,dvipsnames}{xcolor}
\documentclass[sigconf,nonacm, screen]{acmart}
\usepackage{amssymb}
\usepackage{amsmath} 
\usepackage{booktabs} 
\usepackage{array}
\usepackage{graphicx} 
\usepackage{multirow}
\usepackage{color}
\usepackage[table]{xcolor}
\usepackage{makecell} 
\usepackage{float}
\usepackage{subcaption}
\usepackage{colortbl}
\usepackage{placeins}
\usepackage{comment}
\usepackage{pifont}

\usepackage[linesnumbered,vlined,ruled,noend]{algorithm2e}
\usepackage{algorithm2e}
\usepackage{amssymb}
\usepackage{adjustbox}
\usepackage[dvipsnames]{xcolor}
\usepackage{graphicx}

\AtBeginDocument{%
  }

\begin{document}

\title{LongVidSearch: An Agentic Benchmark for Multi-hop Evidence Retrieval Planning in Long Videos}

\author{Rongyi Yu$^*$}
\affiliation{%
  \institution{Harbin Institute of Technology}
  \city{Harbin}
  \country{China}}
\email{2022112619@stu.hit.edu.cn}

\author{Chenyuan Duan$^*$}
\affiliation{%
  \institution{Sun Yat-sen University}
  \city{GuangZhou}
  \country{China}}
\email{duanchy3@mail2.sysu.edu.cn}

\author{Wentao Zhang}
\affiliation{%
  \institution{Peking University \& Zhongguancun Academy}
  \city{Beijing}
  \country{China}}
\email{wentao.zhang@pku.edu.cn}


\thanks{%
$*$Equal contribution.
}

\begin{abstract}
Long video question answering (Long-Video QA) increasingly relies on agentic tool use to retrieve evidence from long videos. In realistic settings, this process often requires multi-hop retrieval, where agents must iteratively gather multiple discontinuous evidence clips. However, existing long-video benchmarks are largely static: they rarely enforce strict multi-hop retrieval and typically lack a standardized evidence-access interface, making it difficult to separate failures in retrieval planning from those in answer generation. To address this gap, we introduce LongVidSearch, a benchmark for evaluating agentic multi-hop evidence retrieval planning in long videos under standardized access constraints. LongVidSearch enforces retrieval necessity: a Hop-$k$ question requires exactly $k$ necessary evidence clips, and removing any single clip renders the question unsolvable. The benchmark contains 3,000 questions over 447 long videos (average length 26 minutes), covering four reasoning categories—State Mutation, Causal Inference, Global Summary, and Visual Tracking—with 2-, 3-, and 4-hop evidence requirements. To ensure fair and controlled evaluation, all agents interact with LongVidSearch through a unified tool interface, which fixes the retrieval backend and isolates the agent’s ability to formulate queries and plan iterative retrieval. In addition to answer accuracy, we measure tool-call cost to analyze the accuracy–efficiency trade-off under identical access conditions. We evaluate VideoAgent-style QA agents with multiple backbone LLMs using three-judge majority voting. GPT-5 achieves the highest accuracy (42.43), outperforming Gemini 3 Pro (30.97) and GPT-4o (19.20), yet remaining below 50\%, highlighting the difficulty of multi-hop retrieval planning. With gold evidence clips, performance becomes near-perfect, confirming retrieval planning as the primary bottleneck. Our code and data are available at \url{https://github.com/yrywill/LongVidSearch}.
\end{abstract}



\maketitle
\vspace{-2mm}
\section{Introduction}

Multimodal foundation models have undergone rapid evolution in recent years~\cite{achiam2023gpt, team2023gemini, bai2025qwen3, an2026genius, an2025unictokens, guo2025video, lin2025perceiveanythingrecognizeexplain, an2024mc, luo2024llm, zhou2024mathscape}. 
From early vision-language alignment models to large-scale multimodal language models (MLLMs), systems such as GPT-4V~\cite{achiam2023gpt}, Gemini~\cite{team2023gemini}, and Qwen-VL~\cite{bai2025qwen3} have demonstrated strong capabilities in visual understanding, cross-modal reasoning, and instruction following. 
More recently, video large language models (Video-LLMs) have extended these advances from static images to dynamic temporal content, enabling long-context video understanding and reasoning across minutes or even hours of footage. 

Despite this progress, a fundamental challenge remains: moving from perception to \textit{deep research}. 
Long-form video content—ranging from surveillance archives and lectures to documentaries and instructional tutorials—has become a primary medium for knowledge. 
Unlike short clips where semantics are often localized within a single shot, long videos distribute relevant information across temporally distant segments. 
Answering a complex query frequently requires identifying multiple discontinuous evidence clips and composing them into a coherent reasoning chain. 
This demands not just recognition, but \textbf{Multi-Hop Search}~\cite{tang2024multihopragbenchmarkingretrievalaugmentedgeneration}: an agent must actively retrieve, verify, and aggregate evidence across time.

However, current evaluation benchmarks do not sufficiently measure this capability. 
Recent long-video benchmarks such as LVBench~\cite{wang2024lvbench} and MLVU~\cite{zhou2024mlvu} have extended video duration and task diversity. 
Yet most adopt static, one-shot evaluation protocols—multiple choice or direct generation—where the model receives a fixed input and produces an answer without explicit retrieval planning. 
These settings do not require iterative evidence acquisition, subgoal decomposition, or adaptive query reformulation. 
As a result, the field lacks a benchmark specifically designed to evaluate \textbf{retrieval-grounded, multi-hop reasoning in long-form videos} under an agentic paradigm.

This gap manifests in two critical deficiencies:

\begin{enumerate}
    \item \textbf{The Necessity Gap (Shortcut Learning).} 
    Many questions labeled as “multi-hop” can still be solved using single-moment visual cues or language priors~\cite{geirhos2020shortcut}, without actually retrieving temporally distributed evidence. 
    Without enforcing \textit{retrieval necessity}, benchmarks risk measuring shortcut exploitation rather than genuine reasoning.

    \item \textbf{The Interaction Gap (Static vs. Agentic Evaluation).} 
    One-shot protocols fail to assess autonomous search behavior. 
    They do not evaluate whether an agent can decompose a problem into subgoals, generate intermediate queries, adaptively call tools, or decide when sufficient evidence has been gathered in long-horizon reasoning.
\end{enumerate}
\begin{figure*}[t]
  \centering
  \includegraphics[width=0.96\textwidth]{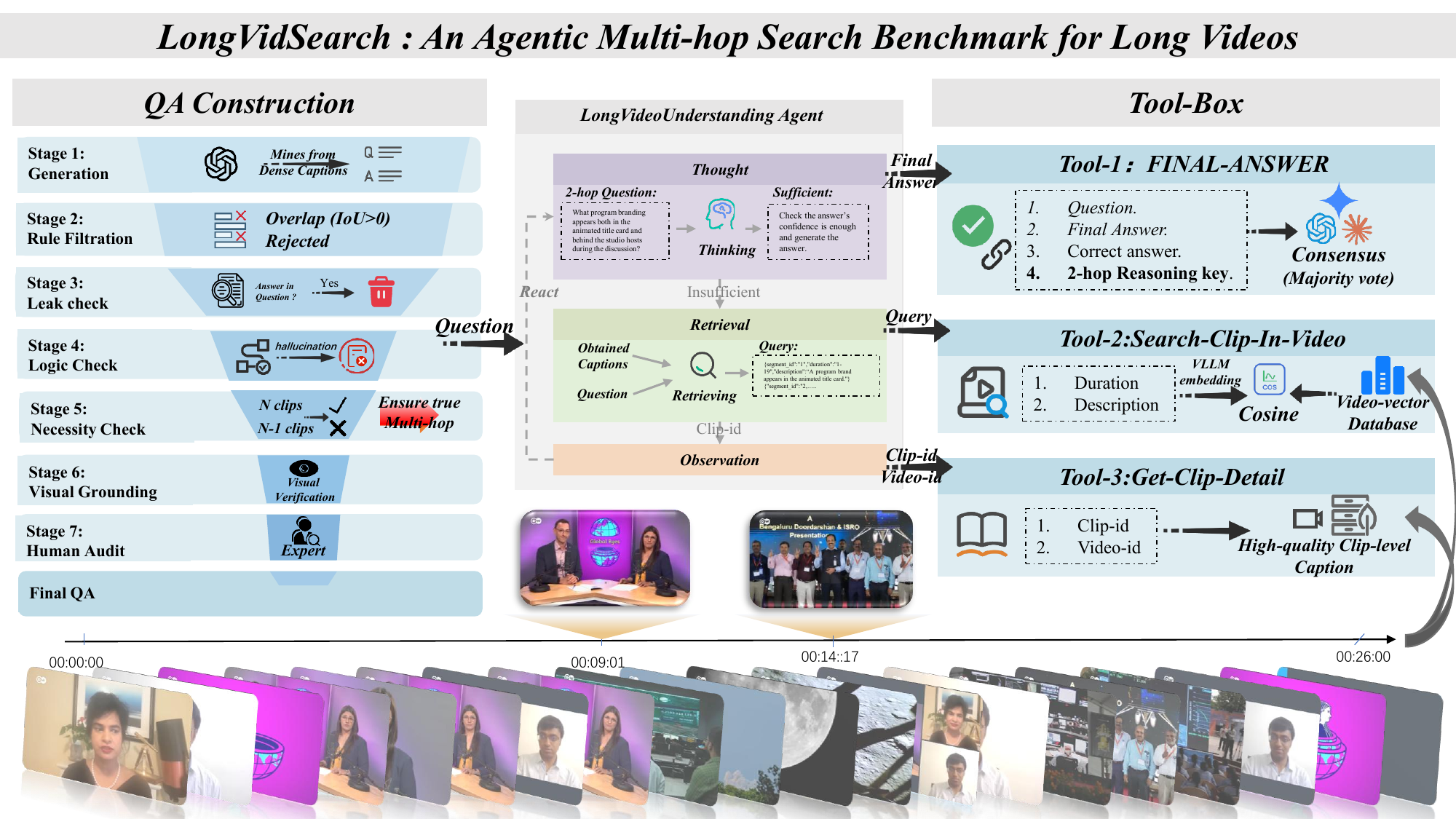}
  \caption{Overview of \textsc{LongVidSearch}. We illustrate the end-to-end pipeline with a representative 2-hop example: the agent iteratively retrieves candidate clips, accesses evidence via standardized tools, and produces a final answer that is scored by a three-judge protocol with majority vote.}
  \label{fig:benchmark_framework}
\end{figure*}
To address these limitations, we introduce \textsc{LongVidSearch}, a benchmark explicitly designed to evaluate \textit{Agentic Search} and \textit{True Multi-Hop Reasoning} in long-form videos. 
Built upon the temporally extensive LoVR dataset~\cite{cai2025lovr}, \textsc{LongVidSearch} operationalizes multi-hop reasoning through the principle of \textbf{retrieval necessity}. 
We define \textbf{Hop-$k$} such that answering the question requires retrieving \textbf{$k$ necessary evidence clips}; removing any single clip renders the question unsolvable.


Constructing such strictly retrieval-necessary questions is non-trivial. 
We therefore propose an \textbf{Agentic Construction Pipeline} that frames data generation as an adversarial process. 
Our pipeline includes: (1)\textbf{Semantic Leakage Auditor} to eliminate tautological or overly localizable queries. (2)\textbf{Temporal Discontinuity Check} to ensure evidence spans non-adjacent segments. (3)\textbf{N-1 Adversarial Ablation Check}: for each candidate question, a verifier agent attempts to answer it while systematically masking one evidence clip at a time. If the question remains solvable under any missing-evidence condition, it is discarded as a shortcut.

Notably, over 45\% of logically valid candidates were filtered out by the necessity check alone, highlighting the prevalence of pseudo-multi-hop samples in automated generation. 
The remaining dataset required minimal human correction, demonstrating the robustness of our adversarial filtering strategy.

Our contributions are summarized as follows:

\begin{itemize}
    \item \textbf{The \textsc{LongVidSearch} Benchmark.} 
    We release \textsc{LongVidSearch}, consisting of \textbf{3,000} QA pairs derived from 447 long-form videos. 
    Questions are explicitly stratified into \textbf{Hop-2 / Hop-3 / Hop-4} based on retrieval necessity, enabling controlled analysis of long-horizon reasoning depth.

    \item \textbf{An Agentic Construction Pipeline.} 
    We establish a scalable methodology for building anti-shortcut benchmarks through adversarial filtering and necessity verification.

    \item \textbf{Standardized Tool-Use Evaluation.} 
    We introduce a unified video-retrieval tool interface, allowing systematic analysis of agentic planning behavior and the accuracy–efficiency trade-off in long-video reasoning.
\end{itemize}

\section{Related Work}

\subsection{Long-Video Understanding Benchmarks}
Recent benchmarks such as LVBench~\cite{wang2024lvbench}, MLVU~\cite{zhou2024mlvu}, Video-MME~\cite{fu2025video} and EgoSchema~\cite{mangalam2023egoschemadiagnosticbenchmarklongform} extend video duration and task coverage, advancing long-context video understanding.
However, they are typically evaluated in \textit{static, one-shot} settings with fixed inputs (e.g., packaged frames/captions), without controlling how evidence is accessed.
This obscures \textbf{agentic} behaviors such as iterative search, planning, and stopping.
\textsc{LongVidSearch} instead evaluates proactive evidence acquisition in long videos through a standardized tool interface.

\subsection{Multi-Hop Retrieval and Necessity Verification}
Multi-hop retrieval is central to text QA (e.g., HotpotQA~\cite{yang2018hotpotqa}, 2WikiMultiHopQA~\cite{ho-etal-2020-constructing}, MuSiQue~\cite{trivedi2021musique}), where answering requires aggregating multiple evidence sources.
We transfer this paradigm to long videos by grounding each question in temporally-discontinuous evidence clips and scaling reasoning depth to \textbf{Hop-2/3/4}, where \textbf{Hop-$k$} denotes \textbf{$k$ necessary evidence clips}.
To reduce shortcut solving~\cite{geirhos2020shortcut} (e.g., single-frame bias~\cite{agrawal2018don}), we enforce retrieval necessity via an \textbf{N-1 adversarial ablation check}: we keep a question only if removing any one evidence clip makes it underdetermined, following evidence-based verification protocols in fact-checking~\cite{thorne2018feverlargescaledatasetfact}.

\subsection{Tool-Augmented Video Agents}
Tool-augmented agents such as VideoAgent~\cite{videoagent}, VideoExplorer~\cite{yuan2025thinkvideosagenticlongvideo}, Deep Video Discovery~\cite{zhang2025deep} and Ego-R1~\cite{tian2025egor1chainoftoolthoughtultralongegocentric} enable iterative search-and-reason pipelines~\cite{yao2022react} for long-video QA.
Yet prior evaluations often use heterogeneous datasets and non-unified tool settings, confounding gains from retrieval backends, tool budgets, and agent policies.
\textsc{LongVidSearch} provides a reproducible testbed with a unified retrieval interface and controlled tool budgets, reporting both \textit{LLM-judged accuracy} and \textit{tool-call cost} to study the accuracy--efficiency trade-off.

\subsection{Data Synthesis}
Recently, data synthesis has emerged as an important technique for improving the performance of large language models (LLMs)~\cite{JCST-2509-15948, bai2024survey}. Prior work has extensively explored data synthesis for both textual and multimodal domains. In the text domain, LLM-driven data synthesis pipelines are typically constructed using complex, workflow-based systems such as DataFlow~\cite{liang2025dataflow, cai2025text2sql, shen2025let, zheng2024pas, liang2024synth}, enabling high-quality synthetic data generation and achieving strong performance across a wide range of downstream tasks. 

In the multimodal domain, data synthesis has also proven effective. For example, prior studies synthesize large-scale image caption datasets~\cite{liu2024synthvlm} or multimodal verification trajectories~\cite{sun2025mm} to enhance the training and reasoning capabilities of vision-language models. In this paper, we follow prior work and extend data synthesis to the video domain.

\section{The \textsc{LongVidSearch} Dataset Construction}

We introduce \textsc{LongVidSearch}, a benchmark constructed to enforce retrieval necessity and visual faithfulness in long video understanding. Departing from the ``quantity-over-quality'' paradigm, we implement a rigorous \textit{Agentic Construction Pipeline} serving as an adversarial funnel. This process systematically filters raw generations through syntactic, semantic, visual, and human validation.

\subsection{Data Source: The LoVR Dataset}
The upper bound of complexity of any VideoQA benchmark is determined by the spatiotemporal richness of its source corpus. We built our benchmark on the \textbf{LoVR dataset (Long Video Retrieval)} ~\cite{cai2025lovr}. We utilize 467 videos with an average duration of \textbf{26 minutes}. This massive temporal window provides a sufficient search space to construct multi-hop queries that span widely separated timestamps, forcing models to utilize long-term memory. We leverage the raw video files and their dense, human-verified captions as the semantic source for question generation.

\begin{table}[t]
  \caption{Statistical Distribution of \textsc{LongVidSearch}. We report the number of questions for each task type stratified by reasoning hops. The \textbf{Total} column indicates the overall prevalence of each task, while the hop-level breakdown reveals the structural complexity inherent to each logic type.}
  \label{tab:stats_distribution}
  \centering
  \small
  \setlength{\tabcolsep}{7pt}
  \renewcommand{\arraystretch}{1.22}
  \begin{tabular}{l c c c | c}
    \toprule
    \textbf{Task Category} & \textbf{2-Hop} & \textbf{3-Hop} & \textbf{4-Hop} & \textbf{Total (Ratio)} \\
    \midrule
    Causal Inference & 436 & 282 & 144 & \textbf{862} (28.7\%) \\
    Global Summary   & 512 & 181 & 166 & \textbf{859} (28.6\%) \\
    Visual Tracking  & 653 & 136 & 61  & \textbf{850} (28.3\%) \\
    State Mutation   & 238 & 119 & 72  & \textbf{429} (14.3\%) \\
    \midrule
    \textbf{Overall Count} & \textbf{1,839} & \textbf{718} & \textbf{443} & \textbf{3,000} \\
    \textit{Overall Percentage} & \textit{61.3\%} & \textit{23.9\%} & \textit{14.8\%} & \textit{100.0\%} \\
    \bottomrule
  \end{tabular}
\end{table}

\begin{table*}[t]
\centering
\small
\setlength{\tabcolsep}{6pt}
\renewcommand{\arraystretch}{1.32}

\caption{General (majority-vote) accuracy (\%) by model, category, and hop level.}
\vspace{-3mm}
\label{tab:acc_model_by_category_byhop}
\begin{tabular}{l c ccc ccc ccc ccc}
\toprule
\multirow{3}{*}{\textbf{Agent Backbone}}
& \multirow{3}{*}{\textbf{Acc(All)}}
& \multicolumn{3}{c}{\textbf{State\_Mutation}}
& \multicolumn{3}{c}{\textbf{Causal\_Inference}}
& \multicolumn{3}{c}{\textbf{Global\_Summary}}
& \multicolumn{3}{c}{\textbf{Visual\_Tracking}} \\
\cmidrule(lr){3-5}\cmidrule(lr){6-8}\cmidrule(lr){9-11}\cmidrule(lr){12-14}
&
& \textbf{2-hop} & \textbf{3-hop} & \textbf{4-hop}
& \textbf{2-hop} & \textbf{3-hop} & \textbf{4-hop}
& \textbf{2-hop} & \textbf{3-hop} & \textbf{4-hop}
& \textbf{2-hop} & \textbf{3-hop} & \textbf{4-hop} \\
\midrule

\rowcolor{blue!15}
\multicolumn{14}{c}{\textbf{Close-Sourced LLMs}} \\

GPT-5
& \cellcolor{gray!15}\textbf{42.43}
& \cellcolor{gray!15}\textbf{38.24} & \cellcolor{gray!15}\textbf{36.13} & \cellcolor{gray!15}\textbf{22.22}
& \cellcolor{gray!15}\textbf{47.71} & \cellcolor{gray!15}\textbf{43.97} & \cellcolor{gray!15}\textbf{39.58}
& \cellcolor{gray!15}\textbf{44.34} & \cellcolor{gray!15}\textbf{35.36} & \cellcolor{gray!15}\textbf{29.52}
& \cellcolor{gray!15}\textbf{49.77} & \cellcolor{gray!15}\textbf{37.50} & \cellcolor{gray!15}\textbf{29.51} \\

Gemini 3 Pro
& 30.97
& 30.25 & 18.49 & 12.50
& 34.17 & 20.92 & 17.36
& 36.72 & 20.44 & 15.66
& 45.48 & 25.00 & 18.03 \\

GPT-4o
& 19.20
& 15.55 & 14.29 & 12.50
& 20.18 & 12.77 & 11.81
& 19.73 & 13.81 & 11.45
& 29.40 & 20.59 & 11.48 \\

GPT-4-mini
& 18.27
& 15.97 & 5.93  & 4.17
& 15.14 & 10.99 & 6.25
& 20.31 & 16.02 & 12.65
& 31.35 & 20.59 & 11.48 \\
\midrule

\rowcolor{green!15}
\multicolumn{14}{c}{\textbf{Open-Sourced LLMs}} \\

Qwen3-VL-32B
& \cellcolor{gray!15}\textbf{29.59}
& \cellcolor{gray!15}\textbf{29.74} & \cellcolor{gray!15}\textbf{27.97} & \cellcolor{gray!15}\textbf{15.49}
& \cellcolor{gray!15}\textbf{29.26} & \cellcolor{gray!15}\textbf{22.86} & \cellcolor{gray!15}\textbf{18.44}
& \cellcolor{gray!15}\textbf{34.19} & \cellcolor{gray!15}\textbf{20.99} & \cellcolor{gray!15}\textbf{16.46}
& \cellcolor{gray!15}\textbf{40.43} & \cellcolor{gray!15}\textbf{25.93} & \cellcolor{gray!15}\textbf{22.95} \\

Qwen3-VL-8B
& 18.58
& 16.67 & 12.71 & 9.72
& 14.81 & 11.43 & 11.19
& 20.59 & 16.67 & 15.34
& 28.84 & 17.78 & 15.25 \\

Qwen2.5-VL-72B
& 25.30
& 23.95 & 17.65 & 12.50
& 26.38 & 21.99 & 15.97
& 29.49 & 20.44 & 15.06
& 34.00 & 22.79 & 9.84 \\

Qwen2.5-VL-7B
& 10.41
& 7.73 & 7.69 & 4.35
& 7.64 & 5.05 & 2.82
& 13.83 & 7.39 & 5.00
& 18.50 & 10.29 & 4.92 \\

Qwen2.5-7B
& 11.10
& 10.92 & 4.20 & 4.17
& 8.72 & 5.32 & 4.86
& 15.82 & 7.73 & 3.61
& 18.99 & 7.35 & 6.56 \\

Llama-3-8B
& 7.73
& 6.72 & 5.88 & 1.39
& 7.57 & 4.96 & 4.86
& 8.20 & 6.08 & 4.22
& 12.71 & 7.35 & 1.64 \\
\bottomrule
\end{tabular}
\end{table*}

\subsection{Taxonomy of Multi-Hop Retrieval Tasks}
\label{sec:taxonomy}

We formalize the \textit{Multi-Hop Search} task as retrieving a set of non-contiguous temporal slices $E = \{s_i, s_j, ...\}$ to resolve a query $Q$. To avoid arbitrary categorization, we derive our taxonomy from two fundamental dimensions: \textbf{(1) Semantic Granularity} (Fine-grained Entity vs.\ Coarse-grained Narrative) and \textbf{(2) Reasoning Paradigm} (Aggregation vs.\ Transition). 
Together, these axes induce four canonical reasoning structures that cover the primary retrieval demands in long-form video question answering. 
We acknowledge that real-world queries are often compositional (e.g., a causal query may implicitly require tracking entities across time). 
However, to maintain a rigorous and unambiguous evaluation protocol, we assign each sample to a single category according to its \textbf{primary reasoning bottleneck}—the dominant logical operation that most directly determines retrieval success. Concretely, we cover:

\textbf{Visual Tracking (Entity + Aggregation)}
This category focuses on \textit{aggregating evidence} for identity persistence. It requires retrieving multiple appearances of an object across long gaps, occlusions, or viewpoint shifts. The primary bottleneck is \textbf{long-term re-identification (ReID)}—ensuring retrieved clips refer to the same entity.

\textbf{State Mutation (Entity + Transition)} 
This category targets \textit{state transitions} of specific objects. Unlike visual tracking (which assumes constancy), it requires retrieving temporally distant segments to contrast the same entity’s attributes (e.g., \textit{intact} vs.\ \textit{broken}). The primary bottleneck is locating the critical \textbf{transition point} in an object’s lifecycle.

\textbf{Causal Inference (Narrative + Transition)}
This category targets \textit{event-to-event transitions}. It requires reasoning from a cause at $t_1$ to an effect at $t_2$. The primary bottleneck is establishing a \textbf{semantic bridge} between two dependent events, understanding the narrative progression rather than just visual matching.

\textbf{Global Summary (Narrative + Aggregation)}
This category is the most abstract. It requires \textit{aggregating} dispersed evidence from $N$ slices to form a holistic conclusion. The primary bottleneck is \textbf{information synthesis}—integrating fragmented narrative clues into a coherent global understanding.

\subsection{The Agentic Construction Pipeline}
\label{sec:pipeline}

We construct \textsc{LongVidSearch} via a seven-stage ``Coarse-to-Fine'' pipeline that functions as an adversarial funnel, successfully distilling 11,612 raw generations into 3,000 high-quality instances. 
Initially, we employ \textbf{GPT-5.2} as a Generator Agent to mine latent reasoning chains from dense video captions, prioritizing recall to cover diverse reasoning types. To ensure the validity of multi-hop retrieval, we apply strict \textbf{Temporal Discontinuity Checks} to discard overlapping evidence clips and a \textbf{Tautology Filter} (Semantic Leakage Auditor) to eliminate questions where the answer is inadvertently self-contained. The tautology check alone removed 33.7\% of the raw data, highlighting the prevalence of semantic leakage in LLM generations.

To strictly enforce retrieval necessity, we introduce a novel \textbf{Adversarial Ablation Protocol (N-1 Check)}. 
Existing benchmarks often suffer from ``shortcut learning,'' where questions are solvable via partial evidence. We address this by challenging a Verifier Agent (\textbf{GPT-5}) to answer a $k$-hop question while systematically masking exactly one necessary evidence clip. A question is retained \textbf{if and only if} the agent returns ``INSUFFICIENT'' for all ablation tests. This rigorous mechanism rejected \textbf{46.0\%} of logically sound candidates, demonstrating that nearly half of conventional multi-hop questions fail to meet strict retrieval necessity standards.

Finally, we bridge the gap between textual captions and visual reality. A \textbf{Qwen3-VL-235B}\cite{bai2025qwen3vltechnicalreport} based Visual Agent verifies pixel-level consistency (e.g., color, action details) to correct caption-induced hallucinations. The pipeline concludes with a human-in-the-loop \textbf{``Adversarial Falsification''} audit, where qualified experts aggressively search for residual loopholes. This multi-layered filtration resulted in a final pass rate of approximately 90\% during the expert review, ensuring the benchmark's high fidelity. Detailed definitions and protocols for each stage are provided in Appendix~\ref{app:construction_details}.

\subsection{LongVidSearch Statistics}
\label{sec:statistics}

The final \textsc{LongVidSearch} benchmark comprises \textbf{3,000 QA pairs} derived from \textbf{447 long-form videos}, with an average duration of approximately \textbf{26 minutes}. The dataset is characterized by its rigorous stratification across reasoning depths and logical categories, ensuring a comprehensive evaluation of agentic capabilities.

\textbf{Complexity Stratification (Hop-level Distribution)}
Unlike benchmarks dominated by single-step retrieval, \textsc{LongVidSearch} is explicitly structured to evaluate multi-step reasoning chains. As detailed in the bottom row of Table~\ref{tab:stats_distribution}, the dataset follows a difficulty gradient: \textbf{61.3\%} of the queries require 2-hop retrieval, serving as a fundamental test of temporal association; \textbf{23.9\%} scale to 3-hop reasoning, testing intermediate memory retention; and \textbf{14.8\%} involve complex 4-hop aggregation, challenging the agent's long-horizon planning abilities.

\textbf{Task Diversity and Distribution}
Table~\ref{tab:stats_distribution} presents the cross-distribution of logical tasks and reasoning depths. To prevent models from overfitting, we maintain a balanced distribution across the four cognitive dimensions. \textbf{Causal Inference (28.7\%)} and \textbf{Global Summary (28.6\%)} constitute the majority, demanding high-level narrative understanding. Notably, Causal Inference exhibits a deeper reasoning structure, with approximately \textbf{50\%} of its questions requiring 3-hop or 4-hop retrieval. In contrast, \textbf{Visual Tracking (28.3\%)} is predominantly 2-hop (76.8\%), reflecting the nature of direct entity re-identification. \textbf{State Mutation (14.3\%)} complements these with fine-grained change detection. This structural diversity ensures that a high score reflects holistic mastery rather than a bias towards specific logic types.
\subsection{Tools}
Our goal is to evaluate \emph{agentic evidence acquisition} under controlled and reproducible evidence access.
To this end, we design a minimal yet complete tool box that (i) \emph{standardizes} the evidence-access process across agents, (ii) \emph{fixes} the retrieval backend to avoid confounding improvements from a stronger retriever, and (iii) \emph{records} tool usage for measuring efficiency.
Concretely, the tools decompose the agent workflow into three atomic operations---\textit{search}, \textit{read}, and \textit{finalize}---so that end-to-end differences primarily reflect the agent's ability to formulate effective queries and plan multi-step retrieval, rather than differences in interface or privileged access to evidence.

Our benchmark provides a standardized tool-calling interface:

\textbf{Search\_Clips\_In\_Video(video\_id, query, top\_k)} retrieves the top-$K$ most relevant clips for a given textual query within the specified video. This tool fixes the retrieval backend for all agents, so performance differences primarily reflect the agent's ability to generate effective queries and plan multi-step retrieval.

\textbf{Get\_Clip\_Detail(clip\_id)} returns a high-quality textual caption for the queried clip, which serves as the contextual evidence for reasoning and answering.

\textbf{FINAL\_ANSWER(answer\_text, evidence\_clip\_ids)} submits the final answer together with the list of viewed evidence clip IDs. The evaluator then computes \textit{Answer Accuracy} and aggregates the \textit{Retrieval Cost} from the tool logs.

\section{Experiments}
\label{sec:experiments}

\subsection{Experimental Settings}
\label{sec:exp_settings}

\subsubsection{Evaluation Metrics.}
We report two metrics to evaluate both answer quality and agentic efficiency under a standardized tool-augmented inference setting.

\textbf{Answer Accuracy.}
We measure accuracy (Acc) by checking whether the predicted answer matches the reference.
When an unambiguous canonical form exists, we apply exact string matching.
For open-form answers where exact matching is insufficient, we adopt an \emph{LLM-as-a-judge}\cite{gu2025surveyllmasajudge}protocol that compares the prediction against the reference under a strict non-hallucination rubric and outputs a binary correctness label.
To reduce evaluator bias, we use three strong LLM judges (GPT-5\cite{singh2025openaigpt5card}, Gemini 3 Pro, and GPT-4o\cite{openai2024gpt4ocard}) and aggregate the final decision via majority voting, reported as \textbf{General}. 
We validate the stability of this protocol in (\S\ref{sec:eval_stability}).

\textbf{Retrieval Cost.}
We quantify inference-time tool usage by counting the number of standardized tool invocations per question.
In our interface, each invocation corresponds to retrieving a candidate clip and then optionally reading its caption; thus, the total number of tool calls directly measures the evidence-access overhead during agentic inference.

\subsubsection{Baselines.}
To benchmark performance under controlled evidence access, we evaluate a VideoAgent-style tool-augmented QA framework with a \emph{fixed} retrieval interface and backend.
All agents share the same tool set and interact with the same retrieval system, so performance differences primarily reflect an agent’s capability in \emph{generating effective queries} and \emph{planning multi-step tool usage}, rather than advantages from a stronger retriever or privileged access to evidence.
We instantiate the same agent framework with different backbone LLMs, while keeping the prompting template, tool budget rules, and evidence-access procedure identical across models.

Importantly, our oracle experiment with golden evidence clips (\S\ref{sec:golden_clips}) shows that once the agent is provided with the correct clips, all backbones can reliably derive the final answer.
This confirms that \textsc{LongVidSearch} primarily tests \emph{retrieval and evidence acquisition}---i.e., the agent’s ability to formulate queries and locate the right evidence---instead of answer generation from already-correct context.

\subsection{Baseline Performance on Our Benchmark}
\label{sec:baseline}


\begin{table}[t]
\centering
\setlength{\tabcolsep}{7pt}
\renewcommand{\arraystretch}{1.15}
\caption{Average tool use by hop level among all categories.}
\label{tab:avg_tool_use_by_hop_grouped}
\begin{tabular}{l c c c c}
\toprule
\textbf{Agent Backbone} &
\textbf{Overall} &
\textbf{2-Hop} &
\textbf{3-Hop} &
\textbf{4-Hop} \\
\midrule
\rowcolor{blue!12}
\multicolumn{5}{c}{\textbf{Closed-Sourced LLMs}} \\
GPT-5        & 9.62 & 9.28 & 9.89 & 10.58 \\
Gemini 3 Pro & 7.37 & 7.30 & 7.39 & 7.58  \\
GPT-4o       & 8.53 & 8.32 & 8.75 & 9.02  \\
GPT-4-mini   & 6.43 & 6.38 & 6.46 & 6.58  \\
\midrule
\rowcolor{green!12}
\multicolumn{5}{c}{\textbf{Open-Sourced LLMs}} \\
Qwen3-VL-32B    & 8.51 & 8.30 & 8.71 & 9.06 \\
Qwen3-VL-8B     & 8.10 & 7.91 & 8.24 & 8.66 \\
Qwen2.5-VL-72B  & 8.78 & 8.48 & 9.03 & 9.58 \\
Qwen2.5-VL-7B   & 7.21 & 7.17 & 7.24 & 7.36 \\
Qwen2.5-7B      & 8.04 & 7.99 & 8.13 & 8.11 \\
Llama-3-8B      & 7.96 & 7.68 & 8.27 & 8.53 \\
\bottomrule
\end{tabular}%
\end{table}

\textbf{Overall accuracy.}
Table~\ref{tab:acc_model_by_category_byhop} reports end-to-end accuracy under the same tool interface and majority-vote evaluation (\textbf{Acc(All)}).
GPT-5 achieves the best overall performance (\textbf{42.43}), followed by Gemini 3 Pro (\textbf{30.97}) and GPT-4o / GPT-4-mini (\textbf{19.20}/\textbf{18.27}).
Among open-source backbones, Qwen3-VL-32B is the strongest (\textbf{29.59}), outperforming the prior-generation Qwen2.5-VL-72B~\cite{bai2025qwen25vltechnicalreport} (\textbf{25.30}) and smaller open models (e.g., Qwen2.5-7B~\cite{qwen2025qwen25technicalreport} \textbf{11.10}, Llama-3-8B~\cite{grattafiori2024llama3herdmodels} \textbf{7.73}).
Despite the clear ranking, even the best backbone remains below \textbf{50\%}, indicating that \textsc{LongVidSearch} is challenging under standardized, tool-mediated evidence access.

\textbf{Hop-level Analysis (deeper hops are harder).}
To characterize the effect of reasoning depth, Table~\ref{tab:acc_model_by_category_byhop} reports accuracy stratified by hop level (2/3/4-hop) across the four capability categories.
Across backbones, accuracy \textbf{consistently decreases} as hop depth increases under the same fixed tool interface, indicating a clear difficulty gradient from 2-hop to 4-hop.
This monotonic degradation also holds \emph{within each category}.
For GPT-5, \textsc{Visual\_Tracking} drops from \textbf{49.77} (2-hop) to \textbf{37.50} (3-hop) and \textbf{29.51} (4-hop), \textsc{State\_Mutation} decreases from \textbf{38.24} to \textbf{36.13} to \textbf{22.22}, and \textsc{Global\_Summary} declines from \textbf{44.34} to \textbf{35.36} to \textbf{29.52}.
The same pattern is observed for strong open backbones such as Qwen3-VL-32B, e.g., \textsc{Visual\_Tracking} \textbf{40.43}$\rightarrow$\textbf{25.93}$\rightarrow$\textbf{22.95} and \textsc{Global\_Summary} \textbf{34.19}$\rightarrow$\textbf{20.99}$\rightarrow$\textbf{16.46}.


\begin{table}[t]
\centering
\setlength{\tabcolsep}{2.5pt}
\renewcommand{\arraystretch}{1.15}

\caption{Performance comparison between the standard agentic setting (\textbf{Standard}) and the Oracle setting (\textbf{Oracle}). Gap ($\Delta$) means the gap between Standard and Oracle.}

\label{tab:oracle_comparison}

\begin{tabular}{l c c c}
\toprule
\textbf{Agent Backbone} & \textbf{Standard Acc (\%)} & \textbf{Oracle Acc (\%)} & \textbf{Gap ($\Delta$)} \\
\midrule
\rowcolor{blue!10} \multicolumn{4}{c}{\textbf{Closed-Sourced LLMs}} \\
GPT-5 & 42.43 & \textbf{100.00} & 57.57 \\
Gemini 3 Pro & 30.97 & 99.97 & 69.00 \\
GPT-4o & 19.20 & 99.40 & 80.20 \\
GPT-4-mini & 18.27 & 98.73 & 80.46 \\
\midrule
\rowcolor{green!10} \multicolumn{4}{c}{\textbf{Open-Sourced LLMs}} \\
Qwen3-VL-32B & 29.59 & 98.56 & 68.97 \\
Qwen3-VL-8B & 18.58 & 96.90 & 78.32 \\
Qwen2.5-VL-72B & 25.30 & 98.60 & 73.30 \\
Qwen2.5-VL-7B & 10.41 & 97.23 & 86.82 \\
Qwen2.5-7B & 11.10 & 97.33 & 86.23 \\
Llama-3-8B & 7.73 & 96.89 & 89.16 \\
\bottomrule
\end{tabular}
\end{table}

\textbf{Tool-call Cost (deeper hops require more tool use).}
We measure retrieval cost as the number of standardized tool invocations per question.
Table~\ref{tab:avg_tool_use_by_hop_grouped} summarizes hop-wise average tool usage aggregated across all categories under the same fixed interface and retrieval backend (see Appendix~\ref{app:Average retrieval cost} for category-level results).
Tool usage generally increases with hop depth for both closed- and open-source backbones, consistent with the larger evidence demand of longer multi-hop chains (e.g., GPT-5: \textbf{9.28}$\rightarrow$\textbf{9.89}$\rightarrow$\textbf{10.58}; Qwen3-VL-32B: \textbf{8.30}$\rightarrow$\textbf{8.71}$\rightarrow$\textbf{9.06}).
Across models, higher accuracy often comes with higher cost: GPT-5 achieves the best Acc(All) (\textbf{42.43}) with the highest overall tool use (\textbf{9.62}), while Gemini 3 Pro attains \textbf{30.97} with fewer calls (\textbf{7.37}).
However, similar costs can yield very different accuracy (e.g., GPT-4o: \textbf{8.53} cost vs.\ \textbf{19.20} Acc(All); Qwen3-VL-32B: \textbf{8.51} cost vs.\ \textbf{29.59} Acc(All)), indicating that tool-call count alone is not a sufficient proxy for effective retrieval planning.
Overall, \textsc{LongVidSearch} supports joint \emph{accuracy--cost} evaluation, enabling finer-grained efficiency assessment beyond accuracy alone.

\subsection{Reasoning with Golden Clips}
\label{sec:golden_clips}

To isolate retrieval from reasoning, we conduct an oracle-style experiment where the agent is provided with the golden (ground-truth) evidence clips.

As shown in Table~\ref{tab:oracle_comparison}, agents achieve near-perfect accuracy (up to \textbf{100\%}) when restricted to these clips.
This indicates that, given the correct evidence, the remaining reasoning difficulty is minimal under our evaluation protocol.
Therefore, the large gap ($\Delta$) to the full benchmark is primarily attributable to \textbf{retrieval failures and multi-hop retrieval planning}---i.e., formulating effective queries and identifying the correct evidence clips---rather than an inability to answer from the appropriate context.

\begin{figure*}
  \centering
  \includegraphics[width=0.90\textwidth]{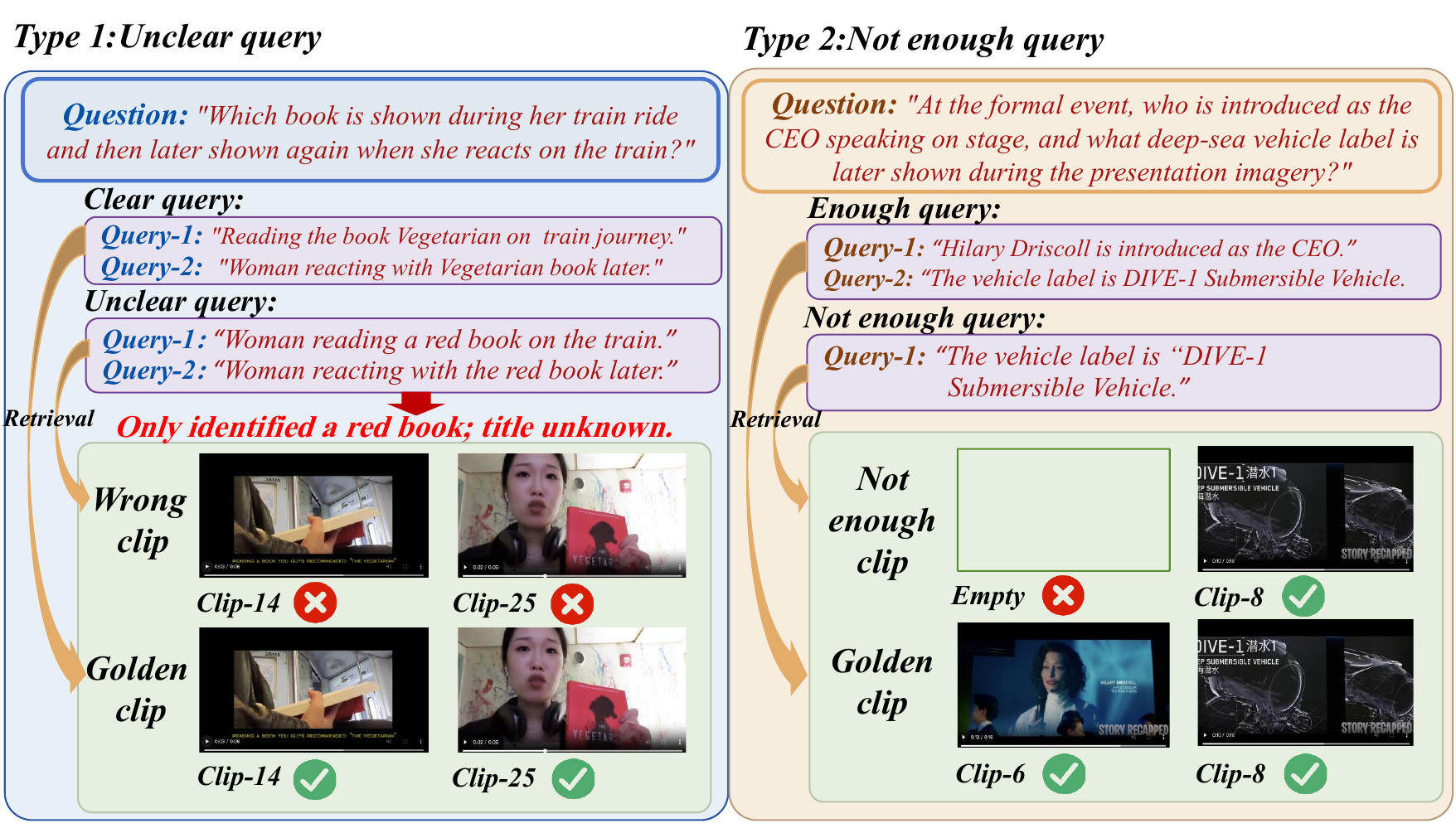}
  \caption{Two types of failure analysis. }
  \label{fig:failure_cases}
\end{figure*}

\subsection{Stability of the Answer Evaluation}
\label{sec:eval_stability}

We adopt a two-stage evaluation protocol to ensure reliable correctness judgments over 3,159 benchmark instances.
For questions with unambiguous references, we first apply exact string matching between the prediction and the ground-truth answer.
For open-form answers where exact match is insufficient, we employ an LLM-as-a-judge procedure.

Importantly, the benchmark provides a \textit{reasoning chain} specifying hop-wise key evidence requirements.
We incorporate these key points into the judging rubric to reduce false positives caused by partial or hallucinated answers.
To further improve robustness, we use a three-judge voting scheme (GPT-5, Gemini 3 Pro, and GPT-4o) and take the majority decision as the final label.

\begin{table}[t]
\centering
\small
\setlength{\tabcolsep}{5.5pt}
\renewcommand{\arraystretch}{1.15}
\caption{Human verification results comparing human final labels with LLM majority-vote labels. \textbf{Disagree rate} is computed as \textit{Disagree / Checked} (\%).}
\label{tab:human_vs_llm}
\begin{tabular}{l c c c}
\toprule
\textbf{Agent Backbone} &
\textbf{Checked (N)} &
\textbf{Disagree (k)} &
\textbf{Disagree rate} \\
\midrule
\rowcolor{blue!12}
\multicolumn{4}{c}{\textbf{Closed-Sourced LLMs}} \\
GPT-5 & 598 & 3 & 0.0050 \\
Gemini 3 Pro & 601 & 5 & 0.0083 \\
GPT-4o & 617 & 6 & 0.0097 \\
GPT-4-mini & 628 & 6 & 0.0096 \\
\midrule
\rowcolor{green!12}
\multicolumn{4}{c}{\textbf{Open-Sourced LLMs}} \\
Qwen3-VL-32B & 607 & 3 & 0.0049 \\
Qwen3-VL-8B & 613 & 6 & 0.0098 \\
Qwen2.5-VL-72B & 620 & 4 & 0.0065 \\
Qwen2.5-VL-7B & 628 & 6 & 0.0096 \\
Qwen2.5-7B & 631 & 7 & 0.0111 \\
Llama-3-8B & 629 & 8 & 0.0127 \\
\midrule
\textbf{Overall} & \textbf{6172} & \textbf{54} & \textbf{0.0087} \\
\bottomrule
\end{tabular}
\end{table}

\textbf{Human verification}
To better validate the reliability of our LLM-judge evaluation, we conduct a human verification study on a subset of instances; details on expert recruitment and training are provided in appendix \ref{app:human_eval} .
We include (i) all cases where the three LLM judges (GPT-5, Gemini 3 Pro, and GPT-4o) do not reach a majority agreement (the \emph{disagreement set}, $N_{\text{disagree}} \approx 200$), and (ii) an additional 400 instances randomly sampled from the majority-agreed pool (the \emph{agreement audit set}).
The audit set is sampled with stratification over categories and hop levels to mitigate coverage bias.
Overall, we verify $N = N_{\text{disagree}} + 400$ instances (about one-fifth of the benchmark).
Across all agent backbones, this yields 6,172 human--LLM label comparisons, and the disagree rate between LLM majority-vote labels and expert decisions is only \textbf{0.0087} (Table~\ref{tab:human_vs_llm}), supporting that our evaluation is stable and reliable.

\subsection{Case Study}
\label{sec:case_study}
We analyze two primary failure modes in \textsc{LongVidSearch}, illustrated in Figure~\ref{fig:failure_cases}.

\textbf{Detail Missing (Unclear Query).}
As shown on the left, the agent successfully locates the target object (a ``red book'') but fails to capture the \textit{specific details} needed to answer. 
The semantic query lacks the precision to trigger text extraction, leaving the specific title \textit{``The Vegetarian''} unresolved despite the correct visual grounding.

\textbf{Broken Evidence Link (Missing Hop).}
As shown on the right, the agent exhibits \textit{Selective Retrieval Failure}. While it successfully retrieves the deep-sea vehicle ``DIVE-1'' (Hop 2), it fails to recall the CEO's identity ``Hilary Driscoll'' (Hop 1).This disconnect prevents the agent from linking the speaker to the presented object, breaking the multi-hop reasoning chain.

\section{Conclusion}
\label{sec:conclusion}

We present \textsc{LongVidSearch}, a retrieval-necessary, evidence-grounded benchmark for \textbf{multi-hop} question answering over long videos, comprising \textbf{3,000 QA pairs} from \textbf{447} videos (avg.\ \textbf{26} minutes) with \textbf{2/3/4-hop} evidence requirements and four capability categories.
By enforcing a \textbf{standardized tool interface} that fixes evidence access and the retrieval backend, \textsc{LongVidSearch} enables controlled evaluation of an agent’s \textbf{query formulation} and \textbf{multi-step evidence acquisition}.
Experiments across both closed- and open-source backbones show that accuracy drops with increasing hop depth, and that reporting both \textbf{accuracy} and \textbf{tool-call cost} reveals a clear accuracy--cost trade-off under identical tool constraints.
An oracle setting with golden clips achieves near-perfect accuracy, confirming \textbf{retrieval} as the primary bottleneck.

We hope \textsc{LongVidSearch} will provide a reliable and reproducible benchmark for evaluating agentic long-video QA, and support future research on \textbf{retrieval planning}, \textbf{evidence-grounded reasoning}, and \textbf{accuracy--cost} trade-offs under standardized tool interfaces.

\clearpage
\bibliographystyle{ACM-Reference-Format}
\bibliography{main}

\appendix


\section{The Agentic Construction Pipeline}
\label{app:construction_details}
Our data construction employs a seven-stage ``Coarse-to-Fine'' pipeline. The adversarial filtration successfully reduced the dataset size from 11,612 raw generations to 3,000 high-quality, retrieval-necessary questions. 

\paragraph{Stage 1: Generation}
We employ \textbf{GPT-5.2} as a \textit{Generator Agent} to mine latent reasoning chains from the dense video captions. By adopting a specific persona, the agent is instructed to identify non-contiguous events that share logical connections and synthesize multi-hop question-answer pairs. This phase prioritizes \textit{recall} over precision, generating a diverse pool of 11,612 raw candidates covering various reasoning types.

\paragraph{Stage 2: Rule-based Filtration}
Following generation, we apply a strict \textbf{Rule-based Filtration} focused on \textbf{Temporal Discontinuity}. A defining characteristic of valid multi-hop retrieval is that evidence segments must be scattered across different temporal moments. We discard candidates where the retrieved evidence clips share temporal overlap (IoU > 0), as these effectively degenerate into single-segment reasoning tasks and violate the fundamental multi-hop constraint. This stage retained 11,423 candidates.
\paragraph{Stage 3: Semantic Leakage Removing (The Tautology Filter)}
A common failure mode in synthetic QA generation is \textit{Answer Leakage}, where the question inadvertently contains the information required to answer it. We deploy a strict Auditor Agent (powered by \textbf{GPT-5}) to identify and discard questions where the answer can be fully inferred or explicitly stated within the question text itself, independent of external context. This rigorous check removed \textbf{33.7\%} of the data, highlighting the prevalence of tautologies in raw LLM outputs.

\paragraph{Stage 4: Logical Solvability Verification}
A Verifier Agent (powered by \textbf{GPT-5}) conducts a Chain-of-Thought (CoT) check to identify internal semantic contradictions. It discards queries where the reasoning chain is broken or relies on hallucinations not present in the source captions, retaining 6,784 robust queries.

\paragraph{Stage 5: Adversarial Necessity Check ($N-1$ Ablation)}
\textbf{This is the core contribution to retrieval rigor.} To eliminate shortcut learning, we implement an \textit{Adversarial Ablation Protocol}. For a question requiring $k$ evidence slices, we mask exactly one slice $s_i$ and challenge a Verifier Agent (\textbf{GPT-5}) to answer using only the remaining context. A question is deemed valid \textbf{if and only if} the agent returns ``INSUFFICIENT'' for all $k$ ablation tests. 


This stage rejected \textbf{46.0\%} of the remaining candidates, proving that nearly half of logically valid ``multi-hop'' questions were not actually retrieval-necessary.

\paragraph{Stage 6: Visual Grounding \& Refinement}
Since textual captions are lossy compressions of reality, we deploy a \textit{Visual Agent} powered by \textbf{Qwen3-VL-235B} to ``watch'' the raw video clips corresponding to the generated timestamps. The agent verifies visual consistency (e.g., color, count, action) and generates a refined answer if discrepancies are found.

\label{qa:example}
\begin{figure*}[!htbp]
\section{Question examples}
\centering
\small
\setlength{\tabcolsep}{6pt}
\renewcommand{\arraystretch}{1.15}

\begin{tabular}{l c p{14cm}}
\toprule
 & \textbf{2-hop} &
\textbf{Question:} \textcolor{NavyBlue}{Which book is shown during her train ride and then later shown again when she reacts on the train?} \\
 & &
\textbf{Answer:} \textcolor{black}{The Vegetarian.} \\
 & &
\textbf{Golden-Clip:} \textcolor{Plum}{[14, 25]} \\
 & &
\textbf{Reasoning-Chain:} Step 1: Slice 14 shows \textcolor{ForestGreen}{the book titled `The Vegetarian' being read on transit.}
Step 2: Slice 25 shows her on a train \textcolor{Maroon}{with a red book and the text about gasping on the train}.
Conclusion: The repeated book is `The Vegetarian.' \\
 & &
\textbf{Category:} \textcolor{BurntOrange}{Visual\_Tracking} \\
 & &
\textbf{Hop-level:} \textcolor{black}{2-Hop} \\
 & &
\textbf{Visual\_proof:} \textcolor{black}{In Clip 1, the text overlay explicitly states...... as she reacts to it.} \\
 & &
\textbf{Logic\_check\_reasoning:} \textcolor{black}{Step 1: Slice 14 explicitly shows a person on a train ...... it again during her reaction.} \\
 & &
\textbf{Video-id:} \textcolor{NavyBlue}{-uvMrMcN0eA} \\
\bottomrule

\vspace{0.6em}


 & \textbf{3-hop} &
\textbf{Question:} \textcolor{NavyBlue}{Across the Italy-focused maps, which trio of cities repeatedly appears as labeled key points (including one that later becomes central to the siege narrative)?} \\
 & &
\textbf{Answer:} \textcolor{black}{Rome, Ravenna, and Naples.} \\
 & &
\textbf{Golden-Clip:} \textcolor{Plum}{[43, 50, 72]} \\
 & &
\textbf{Reasoning-Chain:} Step 1: Slice 43 labels \textcolor{ForestGreen}{Rome, Ravenna, and Naples on a Kingdom of Italy map}.
Step 2: Slice 50 again labels \textcolor{Maroon}{Rome, Ravenna, and Naples among marked cities}.
Step 3: Slice 72 \textcolor{TealBlue}{repeats Rome and Naples prominently and includes Ravenna in the same political landscape}.
Conclusion: The recurring trio is Rome, Ravenna, and Naples. \\
 & &
\textbf{Category:} \textcolor{BurntOrange}{Global\_Summary} \\
 & &
\textbf{Hop-level:} \textcolor{black}{3-Hop} \\
 & &
\textbf{Visual\_proof:} \textcolor{black}{All three clips show a map of the Kingdom of Italy...... supporting the narrative context.} \\
 & &
\textbf{Logic\_check\_reasoning:} \textcolor{black}{Step 1: Slice 43 lists the cities ......shows the trio as labeled cities on Italy-focused maps.} \\
 & &
\textbf{Video-id:} \textcolor{NavyBlue}{-7wwfGJXEZg} \\
\bottomrule

\vspace{0.6em}


 & \textbf{4-hop} &
\textbf{Question:} \textcolor{NavyBlue}{How does her dessert-prep storyline progress from choosing a fruit at the store to a final plated result that includes more than one kind of fruit?} \\
 & &
\textbf{Answer:} \textcolor{black}{She selects strawberries at the store, then later dips strawberries (and includes grapes too) in chocolate, ending with a final results plate containing strawberries and grapes.} \\
 & &
\textbf{Golden-Clip:} \textcolor{Plum}{[35, 54, 59, 60]} \\
 & &
\textbf{Reasoning-Chain:} Step 1: Slice 35 shows \textcolor{ForestGreen}{her choosing strawberries while shopping}.
Step 2: Slice 54 shows \textcolor{Maroon}{grapes being added and the idea of covered fruits}.
Step 3: Slice 59 shows \textcolor{TealBlue}{dipping strawberries into melted chocolate}.
Step 4: Slice 60 shows \textcolor{RoyalPurple}{`final results' with strawberries and grapes on a plate}.
Conclusion: Shopping leads to multi-fruit chocolate prep and a finished mixed-fruit plate. \\
 & &
\textbf{Category:} \textcolor{BurntOrange}{Causal\_Inference} \\
 & &
\textbf{Hop-level:} \textcolor{black}{4-Hop} \\
 & &
\textbf{Visual\_proof:} \textcolor{black}{Clip 1 shows her selecting strawberries...... chocolate-covered strawberries and grapes.} \\
 & &
\textbf{Logic\_check\_reasoning:} \textcolor{black}{Step 1: Slice 35 shows selecting/purchasing strawberries...... to final plated outcome.} \\
 & &
\textbf{Video-id:} \textcolor{NavyBlue}{-uvMrMcN0eA} \\
\bottomrule
\end{tabular}

\caption{Data examples of different hop. Each block lists the question, answer, golden clips, reasoning chain,category,hop-level, video-id and verification fields.}
\label{fig:data_examples_three_blocks}
\end{figure*}

\begin{table*}[t]
\section{Retrieval cost(tool-invocation count) by model, category, and hop level}
\label{app:Average retrieval cost}
\centering
\small
\setlength{\tabcolsep}{6pt}
\renewcommand{\arraystretch}{1.28}
\caption{Average retrieval cost by model, category, and hop level.}
\label{tab:cost_model_by_category_byhop}
\begin{tabular}{l ccc ccc ccc ccc}
\toprule
\multirow{3}{*}{\textbf{Agent Backbone}}
& \multicolumn{3}{c}{\textbf{State\_Mutation}}
& \multicolumn{3}{c}{\textbf{Causal\_Inference}}
& \multicolumn{3}{c}{\textbf{Global\_Summary}}
& \multicolumn{3}{c}{\textbf{Visual\_Tracking}} \\
\cmidrule(lr){2-4}\cmidrule(lr){5-7}\cmidrule(lr){8-10}\cmidrule(lr){11-13}
& \textbf{2-hop} & \textbf{3-hop} & \textbf{4-hop}
& \textbf{2-hop} & \textbf{3-hop} & \textbf{4-hop}
& \textbf{2-hop} & \textbf{3-hop} & \textbf{4-hop}
& \textbf{2-hop} & \textbf{3-hop} & \textbf{4-hop} \\
\midrule
\rowcolor{blue!15}
\multicolumn{13}{c}{\textbf{Close-Sourced LLMs}} \\

GPT-5
& 9.26 & 9.69 & 10.13
& 9.40 & 10.02 & 11.02
& 9.17 & 9.56 & 10.50
& 9.29 & 10.27 & 10.23 \\
Gemini 3 Pro
& 7.35 & 7.16 & 7.29
& 7.21 & 7.33 & 7.62
& 7.37 & 7.39 & 7.52
& 7.29 & 7.76 & 7.98 \\
GPT-4o
& 8.41 & 8.70 & 8.81
& 8.44 & 8.83 & 8.99
& 8.11 & 8.60 & 9.22
& 8.37 & 8.82 & 8.82 \\
GPT-4-mini
& 6.33 & 6.32 & 6.37
& 6.45 & 6.43 & 6.72
& 6.34 & 6.56 & 6.59
& 6.40 & 6.54 & 6.43 \\
\midrule
\rowcolor{green!15}
\multicolumn{13}{c}{\textbf{Open-Sourced LLMs}} \\

Qwen3-VL-32B
& 8.32 & 8.52 & 8.77
& 8.53 & 8.89 & 9.49
& 8.24 & 8.57 & 8.86
& 8.18 & 8.66 & 8.84 \\
Qwen3-VL-8B
& 8.02 & 8.05 & 8.64
& 8.20 & 8.33 & 8.75
& 7.74 & 8.29 & 8.67
& 7.79 & 8.17 & 8.44 \\
Qwen2.5-VL-72B
& 8.68 & 9.01 & 9.32
& 8.50 & 9.14 & 9.74
& 8.27 & 8.77 & 9.58
& 8.53 & 9.21 & 9.48 \\
Qwen2.5-VL-7B
& 7.19 & 7.28 & 7.29
& 7.17 & 7.24 & 7.36
& 7.16 & 7.31 & 7.25
& 7.16 & 7.10 & 7.74 \\
Qwen2.5-7B
& 7.99 & 8.20 & 7.99
& 8.18 & 8.03 & 8.16
& 8.00 & 8.08 & 8.02
& 7.85 & 8.36 & 8.34 \\
Llama-3-8B
& 7.85 & 8.03 & 8.87
& 7.88 & 8.39 & 8.74
& 7.69 & 7.96 & 8.11
& 7.52 & 8.50 & 8.28 \\
\bottomrule
\end{tabular}
\end{table*}

\begin{table*}[t]
\centering
\small
\setlength{\tabcolsep}{6.0pt}
\renewcommand{\arraystretch}{1.18}
\caption{Human evaluation criteria for the multi-hop question cross three dimensions.}
\label{tab:judge_rubric}
\begin{tabular}{l p{0.36\textwidth} p{0.36\textwidth}}
\toprule
\textbf{Criterion} & \textbf{Score 1 (Correct)} & \textbf{Score 0 (Incorrect)} \\
\midrule
\textbf{Answer correctness} &
All required entities/values are correct and unambiguous. &
Any required detail is wrong/missing, or \texttt{insufficient} despite available evidence. \\

\textbf{Hop-wise evidence} &
Satisfies \emph{all} hop key points in the reasoning chain; supported by evidence clips. &
Misses any hop key point, or uses unsupported speculation/hallucination. \\

\textbf{Extra details} &
Extra details are allowed if they do not contradict evidence or change the answer. &
Adds contradictory/fabricated details or changes the answer beyond evidence. \\

\bottomrule
\end{tabular}
\end{table*}

\paragraph{Stage 7: Final Human Audit}
To guarantee benchmark integrity, we implemented a rigorous human-in-the-loop verification protocol acting as the final quality gate.

\noindent\textbf{Annotator Qualification.} We recruited a panel of \textbf{10 postgraduate researchers} specializing in Computer Vision and Multi-modal Learning. Before the formal audit, all annotators underwent a qualification phase, requiring them to pass a screening test with 100 control samples (50 valid, 50 flawed) with an accuracy threshold of 95\% to ensure alignment with our rigorous standards.To prevent cognitive fatigue and ensure high vigilance, we randomly partitioned the 3,392 candidates among the experts, resulting in a manageable workload of \textbf{approximately 300 samples per auditor}.

\noindent\textbf{The Adversarial Review Protocol.} Unlike passive annotation, experts operated under an \textbf{``Adversarial Falsification''} mandate. They were instructed to aggressively search for flaws rather than verify correctness. Using a custom verification interface, auditors watched the raw video segments referenced by the generated timestamps and assessed each QA pair against a strict \textbf{Three-Point Rejection Rubric}:
\begin{enumerate}
    \item \textit{Visual Hallucination:} The reasoning relies on visual details not present or ambiguous in the raw pixel data (e.g., misidentifying a blurry object).
    \item \textit{Logic Loophole:} The reasoning chain contains non-sequiturs or requires external knowledge outside the video context.
    \item \textit{Retrieval Unnecessity:} The question is theoretically solvable via language priors or a single frame, failing the strict multi-hop requirement.
\end{enumerate}
During this exhaustive audit of all 3,392 machine-verified samples, experts \textbf{rejected 392 instances (11.6\%)} containing residual ambiguity and manually \textbf{modified 4 instances ($<$0.1\%)} to refine linguistic precision. The fact that approximately 90\% of the candidates passed this stringent review without rejection provides compelling evidence for the efficacy of our automated pipeline, confirming that the adversarial filters (Stages 3--6) successfully maintained high data purity.

To strictly quantify the final quality, we conducted a post-audit verification by \textbf{randomly sampling 100 QA pairs} for a blind review. The inspection revealed that \textbf{100\% of these sampled instances were error-free}. This perfect validation rate, combined with the high acceptance rate of the full audit, establishes \textsc{LongVidSearch} as a high-fidelity benchmark.

\section{Human Evaluation Details and Ethical Considerations}
\label{app:human_eval}

\textbf{Annotator Recruitment.}
Human evaluators were recruited from adult participants fluent in English.
All annotators were informed of the purpose of the study and participated voluntarily.
To ensure the reliability of human judgments, annotators were required to have prior experience in evidence-based QA or text evaluation tasks and to pass a screening test.
The screening test assesses: (i) the ability to follow hop-wise evidence requirements, (ii) the ability to distinguish grounded answers from speculation/hallucination, and (iii) the correct handling of \texttt{insufficient} cases (answerable vs.\ unanswerable given evidence).


\textbf{Annotation Procedure and Criteria.}
Annotators were presented with a question, the model prediction, the benchmark-provided reasoning chain with hop-wise key points, and the corresponding evidence clips (IDs and captions/frames).
Annotators were asked to assign a binary correctness label following the rubric in Table~\ref{tab:judge_rubric}.
A prediction is labeled as \texttt{Correct} (Score~1) only if it answers the question correctly \emph{and} satisfies \emph{all} hop-wise evidence requirements supported by the evidence clips; otherwise it is labeled as \texttt{Incorrect} (Score~0), including cases where the model outputs \texttt{insufficient} despite the answer being available from evidence.
Detailed guidelines and illustrative examples were provided in advance to reduce ambiguity.

\textbf{Verification Set Construction.}
We verify two complementary subsets to validate the reliability of LLM-judge labels:
(1) \emph{Disagreement set}: all instances where the three LLM judges (GPT-5, Gemini 3 Pro, and GPT-4o) do not reach a majority agreement ($N_{\text{disagree}} \approx 200$ in our experiments), which are adjudicated by domain experts;
(2) \emph{Agreement audit set}: an additional 400 instances sampled from the majority-agreed pool for random auditing.
To mitigate coverage bias, the audit set is selected in a stratified manner across categories and hop levels.
Across all evaluated agent backbones, this protocol yields 6172, human--LLM label comparisons.

\textbf{Human--LLM Judge Consistency.}
We compare the final human labels against the LLM majority-vote labels on all verified instances.
As shown in Table~\ref{tab:human_vs_llm}, the overall mismatch rate is 0.87\% (54/6172), and per-backbone mismatch rates remain around 1\% or lower, indicating that the LLM majority-vote evaluation is stable under our hop-aware rubric.

\textbf{Compensation.}
Annotators were compensated at a fixed rate of \$15 per hour.
The compensation was independent of annotators' ratings to avoid incentive bias, and no performance-based or outcome-dependent rewards were provided.

\textbf{Ethical Considerations.}
The human evaluation process did not involve the collection of any personally identifiable information.
All evaluation content was anonymized and contained no sensitive personal data.
Annotators were informed that the evaluated responses were model-generated and were instructed to focus solely on evidence-grounded correctness.
Given the non-invasive nature of the task and the absence of personal data collection, this study does not raise significant ethical concerns and does not require institutional review board approval, consistent with prior work in NLP and multimodal evaluation.

\end{document}